\documentclass[12pt]{article}

\usepackage{amsmath}
\usepackage{amssymb}

\usepackage{graphicx}

\usepackage{color}
\usepackage{xcolor}
\definecolor{shadecolor}{RGB}{200,200,200}
\usepackage{soul}
\usepackage[framemethod=tikz]{mdframed}

\usepackage{setspace}

\usepackage{indentfirst}

\usepackage[hang,flushmargin]{footmisc}

\usepackage{threeparttable}
\usepackage{threeparttablex}
\usepackage{longtable}
\usepackage{subcaption}
\usepackage{booktabs}

\usepackage{hyperref}

\usepackage{fixltx2e}

\usepackage{bigints}

\usepackage{rotating}

\usepackage[left=1.5in, right=1.0in, top=2.0in, bottom=1.0in, paper=letterpaper, includefoot]{geometry}

\usepackage[labelfont=bf,labelsep=period,justification=raggedright]{caption}

\date{}

\begin{document}
\thispagestyle{empty}

% set line spacing and spaces after period
\singlespacing
\frenchspacing
\pagenumbering{roman}

\newpage
\newgeometry{left=1.5in, right=1.0in, top=1.0in, bottom=1.0in, paper=letterpaper, includefoot}
\pagenumbering{arabic}
\setcounter{page}{1}
\setcounter{table}{0}
\setcounter{figure}{0}
\doublespacing

\vspace*{0.5in}
\begin{center}
{\Large
\textbf{Using SVM to pre-classify government purchases} \\
Thiago Marzag\~{a}o\footnote{Data scientist at Conselho Administrativo de Defesa Econ\^{o}mica.} \\
thiago.marzagao@cade.gov.br
}
\end{center}

\section*{Abstract}

\singlespacing

The Brazilian government often misclassifies the goods it buys. That makes it hard to audit government expenditures. We cannot know whether the price paid for a ballpoint pen (code \#7510) was reasonable if the pen was misclassified as a technical drawing pen (code \# 6675) or as any other good. This paper shows how we can use machine learning to reduce misclassification. I trained a support vector machine (SVM) classifier that takes a product description as input and returns the most likely category codes as output. I trained the classifier using 20 million goods purchased by the Brazilian government between 1999-04-01 and 2015-04-02. In 83.3\% of the cases the correct category code was one of the three most likely category codes identified by the classifier. I used the trained classifier to develop a web app that might help the government reduce misclassification. I open sourced the code on GitHub;  anyone can use and modify it.

\doublespacing

\section*{1. Problem}

The Brazilian government classifies the goods it buys according to a particular taxonomy: the CATMAT (Catalog of Materials). When the government purchases a new good, the person in charge of classifying it must choose one of the 560 CATMAT classes. The cognitive load is heavy, especially if the person is new to CATMAT. The result is misclassification. For instance, of the 30,774 goods classified as wheeled vehicles (CATMAT class \#2320) between 1999-04-01 and 2015-04-02, 17,469 cost less than R\$ 1,000 (US\$ 250 as of 2015-12-01). That price seems too low for a wheeled vehicle. When we inspect the product descriptions we see that the problem is misclassification: spare parts (CATMAT classes \#2510, \#2520, \#2530, and \#2590) are often misclassified as wheeled vehicles (CATMAT class \#2320).

When the misclassification rate is so high it is hard to audit government expenditures. In particular, it is hard to know whether the price paid for a given item was reasonable (i.e., similar to the prices paid for the same item before) or not. Auditing authorities waste resources investigating legitimate purchases, on the one hand, and fail to investigate fraudulent purchases, on the other hand.

\section*{2. Machine-based classification}

\subsection*{2.1 Basic idea}

To reduce misclassification I trained a support vector machine (SVM) classifier that takes a product description as input and returns the three most likely CATMAT classes as output. The idea is to semi-automate the process: we want to suggest likely classes to the person classifying the good (the user of our app). He or she may choose among the suggested classes or ignore them and choose a different class. For instance, say some public hospital is buying insulin syringes. The hospital will describe the product in detail on the invitation to bid - say, as ``disposable insulin syringes with 0.5mL and ultra-thin (6mm) needle, 50 units". The user feeds that description to the SVM classifier, which then suggests three CATMAT classes: \#6515 (medical and surgical equipment and supplies), \#6550 (supplies for medical diagnostic and testing), and \#6640 (laboratory equipment and supplies). The user chooses one of these three classes or, alternatively, discards them and chooses a different class.

We cannot fully automate the process (i.e., substitute the SVM classifier for the human classifier), for two reasons. First, in about 16.7\% of the cases none of the SVM-suggested classes is the correct one (more on this later). Hence a human must remain in charge of the final decision. But in the remaining 83.3\% of the cases the SVM classifier will be correct, which should reduce classification error. And perhaps this is a foundation upon which a fully automated classifier can be created in the future.\footnote{Also, it is possible that 83.3\% of correct classifications is more than what the average human classifier achieves. Recall the example discussed before: over 50\% of what is assigned to CATMAT class \#2320 (wheeled vehicles) should have been assigned to other classes (spare parts: CATMAT classes \#2510, \#2520, \#2530, and \#2590).}

Second, after the user has chosen a CATMAT class he or she must also choose a subclass and an item. The CATMAT is a hierarchical taxonomy: it comprises 79 groups, 560 classes, 18,380 subclasses, and 217,907 items (as of 2015-04-02). But the SVM classifier cannot offer any suggestions when it comes to subclasses or items: with 18,380 or 217,907 categories there is just not enough data to train the classifier; its accuracy would be too low. But reducing misclassification at the CATMAT level already improves the quality of the data, even if the user chooses the wrong subclass or item - research and auditing done at the class level will be more reliable. And perhaps the user is more likely to choose the correct subclass and item when the CATMAT class is correct.

\subsection*{2.2 Support Vector Machines}

SVM is a well-known classification algorithm first introduced by Vapnik and Lerner (1963) and later popularized by Boser, Guyon and Vapnik (1992) and by Cortes and Vapnik (1995). It learns from manually classified samples (the training samples) and then uses that knowledge to classify new samples. Here we have a total of 28,420,761 samples, which are all the goods the Brazilian government purchased between 1999-04-01 and 2015-04-02.\footnote{The data are available at http://dw.comprasnet.gov.br} I split the samples into 70\% training, 15\% validation, and 15\% testing. The classifier ``sees" the description \emph{and} the CATMAT class of each of the training samples and thus learns how to map descriptions onto CATMAT classes.\footnote{The mathematics of SVM are beyond the scope of this paper. For a full mathematical treatment of SVM see Hastie, Tibshirani and Friedman (2009).} For instance, the classifier learns that when ``stapler" appears in the description the good is usually classified under CATMAT class \#7520. It learns that when ``tea" appears in the description the good is usually classified under CATMAT class \#8955. And so on.

Once the classifier is trained we use the validation samples to assess its performance. For each validation sample and each CATMAT class the classifier estimates the probability that the sample belongs to that class. Hence for each validation sample the classifier estimates 560 probabilities.\footnote{This is a departure from `vanilla' SVM, in which the classifier merely assigns a class to each validation sample, without any probability estimates. This departure is necessary because we want to suggest not one but three different CATMAT classes to the person in charge of the classification. This comes at the cost of added mathematical complexity and computational cost: in order to produce the probability estimates we must run logistic regression on the SVM scores and use five-fold cross-validation. For the full mathematical treatment of the matter see Platt (2000) and Wu and Weng (2002).} The estimation is word-based. For instance, having the word ``stapler" in the description increases the probability that the sample belongs to CATMAT class \#7520 and decreases the probability that it belongs to CATMAT class \#8955. When all estimations are done we check how often the CATMAT class of highest probability is the correct one. We then go back to the training phase, changing one or more model parameters (like the learning rate or the regularization term). We assess the classifier's performance once more. We train the model again, with new changes in the parameters. And so on and so forth until we are satisfied with the results.

When the classifier performs well enough with the validation samples we check how well it performs with the testing samples. That is the final result - once the testing samples are used we cannot go back and adjust the parameters again, lest we overfit the classifier (i.e., lest we make the classifier too responsive to the idiosyncrasies of our data and thus non-generalizable).

\section*{3. Data}

Before we can train the SVM classifier we need to collect, clean, and process the data. All the data are in the Comprasnet database.\footnote{http://dw.comprasnet.gov.br} The descriptions\footnote{The description is the concatenation of three separate fields: DS\_ITCP\_COMPL\_ITEM\_COMPRA, DS\_ITCP\_COMPL\_ITEM\_COMPRA\_CONT, and DS\_ITCP\_UNIDADE\_FORNECIMENTO} are in the D\_ITCP\_ITEM\_COMPRA table. The CATMAT classes \footnote{ID\_ITCP\_TP\_COD\_MAT\_SERV} are in the F\_ITEM\_COMPRA table. A common key\footnote{ID\_ITCP\_ITEM\_COMPRA} allows JOINs between these two tables. I discarded the cases whose description and/or class was empty or null we are left. I also discarded the cases that corresponded to services (about 10\% of the total), as services are too unique and idiosyncratic for algorithmic classification. The result was a total of 28,420,761 samples; as mentioned above, they correspond to all the goods the Brazilian government purchased between 1999-04-01 and 2015-04-02.

I lower-cased all the descriptions and removed all special characters, numbers, and one-character words. I also converted plural words to singular words, using the first step of a Portuguese stemmer algorithm (the RSLP algorithm, created by Orengo and Huyck [2001]). Finally, I removed all words that only appeared once in the entire dataset.

I transformed each document (description) into a vector of word counts and merged all vectors. This resulted in a term-frequency matrix - a matrix whose rows represent words, columns represent documents, and each entry is the frequency of word \emph{i} on document \emph{j} (i.e., the term-frequency, $TF_{ij}$).

Next I apply the $TF-IDF$ transformation. The $TF-IDF$ of each entry is given by its term-frequency ($TF_{ij}$) multiplied by $ln(n / df_{i})$, where $n$ is the total number of documents and $df_{i}$ is the number of documents in which word $i$ appears (i.e., the word's document frequency -- $DF$; the $ln(n / df_{i})$ ratio thus gives us the inverse document frequency -- $IDF$).

What the $TF-IDF$ transformation does is increase the importance of the word the more it appears in the document but the less it appears in the whole corpus. Hence it helps us reduce the weights of inane words like `the' (`o', in Portuguese), `of' (`de'), etc and increase the weights of discriminant words (i.e., words that appear a lot but only in a few documents). For more details on $TF-IDF$ see Manning, Raghavan, and Sch\"utze (2008).

The next step is normalization. Here we have documents of widely different sizes, ranging from two or three words to dozens of words. Longer documents contain more unique words and have larger $TF$ values, which may skew the results (Manning, Raghavan, and Sch\"utze 2008). To avoid that we normalize the columns of the $TF-IDF$ matrix, transforming them into unit vectors.

The resulting $TF-IDF$ matrix has dimensions 28,420,761 (descriptions) by 505,938 (unique words). It is this matrix that we use to train, validate, and test the SVM classifier.

\section*{4. Results}

Our SVM classifier achieved an accuracy of 83.3\%.\footnote{The code I used can be found at \url{http://thiagomarzagao.com/papers} See code for details on model parameters. I used the Python packages NLTK (Bird, Loper and Klein 2009) and scikit-learn (Pedregosa et al. 2011).}

The errors seem to have two main causes. First, misclassification in the training data. The SVM classifier is learning from imperfect examples. As we discussed before, half the goods currently classified as wheeled vehicles (CATMAT class \#2320) are actually spare parts (CATMAT classes \#2510, \#2520, \#2530, and \#2590). Second, class frequency. CATMAT classes that are bought more often tend to be more accurately classified by the SVM classifier. This is expected: for these classes there are more training samples, so the classifier has a larger set of descriptions from which to learn. In fact, there is a negative linear correlation between class frequency and misclassification rate (-0.36, with p $<$ 0.0001).\footnote{Misclassification rate = frequency of misclassification for the class / frequency of the class.}

\section*{5. Web app}

I used the trained classifier to build an open source web app. Its interface is very simple: there is a single input form, wherein the user writes or pastes the description of the good being purchased (as it appears on the invitation to bid, for example). The user then clicks 'submit' and is shown the three most likely CATMAT classes.

In the example below we inputed the description of an air conditioning unit and got back three CATMAT classes: \#4120 (air conditioning equipment), with probability 58\%; \#4130 (air conditioning and cooling components), with probability 22\%; and \#6550 (supplies for medical diagnostic and testing), with probability 4\%. Here the first class - the one with the highest probability - is the correct one.

\begin{figure}[h!]
  \includegraphics[width=1.0\textwidth]{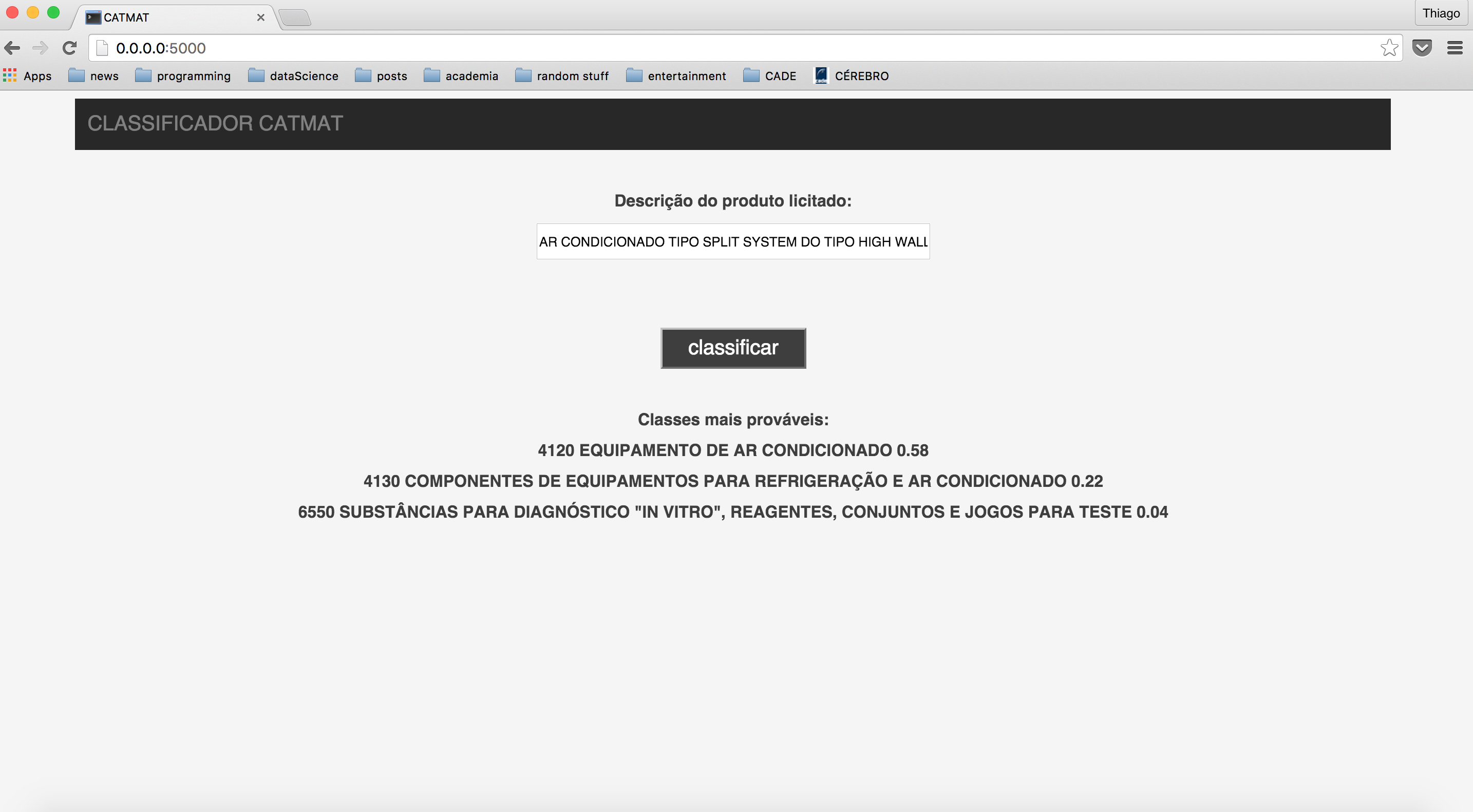}
  \caption{A web app for choosing CATMAT classes}
  \label{figure1}
\end{figure}

Initially the idea was to put the app up online but the cost was prohibitive. The trained classifier is heavy (the coefficients take up over 4GB on disk) and requires a computer with large RAM - at least 16GB but ideally more. On cloud providers like Amazon Web Services and Google Compute Engine that would cost around US\$ 200 per month (R\$ 800 as of 2015-12-01).

The code for the app is available on GitHub though: \url{https://github.com/thiagomarzagao/catmatfinder}. I made the code open source so that anyone can use and modify it.

\newpage
\vspace*{0.5in}

\singlespacing
\begin{center}
\section*{References}
\end{center}

Bird, Steven, Edward Loper, and Ewan Klein. 2009. ``Natural Language Processing with Python." O'Reilly Media Inc.

\vspace*{1\baselineskip}

Boser, Bernhard, Isabelle Guyon and Vladimir Vapnik. 1992. ``A training algorithm for optimal margin classifiers." In D. Haussler, editor, \emph{Proceedings of the Annual Conference on Computational Learning Theory}, 144-152.

\vspace*{1\baselineskip}

Cortes, Corrina and Vladimir Vapnik. 1995. ``Support vector networks." \emph{Machine Learning}, 20: 273-297.

\vspace*{1\baselineskip}

Hastie, Trevor, Robert Tibshirani and Jerome Friedman. 2009. \emph{The elements of statistical learning: data mining, inference, and prediction}. Springer.

\vspace*{1\baselineskip}

Manning, Raghavan, Prabhakar Raghavan, and  Hinrich Sch\"utze. 2008. ``Introduction to information retrieval". Cambridge University.

\vspace*{1\baselineskip}

Orengo, Viviane, and Christian Huyck. ``A stemming algorithmm for the portuguese language." spire. In 8th International Symposium on String Processing and Information Retrieval (SPIRE), IEEE, 183-193.

\vspace*{1\baselineskip}

Pedregosa et al. 2011. ``Scikit-learn: Machine Learning in Python". \emph{Journal of Machine Learning Research}, 12: 2825-2830.

\vspace*{1\baselineskip}

Platt, John. 2000. Probabilistic outputs for support vector machines and comparison to regularized likelihood methods. In A.J. Smola, P.L. Bartlett, B. Sch\"olkopf, and D. Schuurmans, editors, \emph{Advances in Large Margin Classifiers}. MIT Press.

\vspace*{1\baselineskip}

Vapnik, Vladimir and Aleksandr Lerner. 1963.  ``Pattern recognition using generalized portrait method." \emph{Automation and Remote Control}, 24: 774-780.

\vspace*{1\baselineskip}

Wu, Ting-Fan, Chih-Jen Lin and Ruby Weng. 2002. ``Probability estimates for multi-class classification by pairwise coupling." Journal of Machine Learning Research, 2: 615-637.

\end{document}